# Bearings Fault Detection Using Hidden Markov Models and Principal Component Analysis Enhanced Features

Akthem Rehab[1], Islam Ali[1,2], Walid Gomaa[3], and M. Nashat Fors[1]

[1]*Production Engineering Department, Alexandria University, Alexandria, Egypt*
Akthem.Rehab@alexu.edu.eg        Nashat.Fors@alexu.edu.eg
[2]*Department of Industrial and Manufacturing Engineering, Egypt-Japan University of Science and Technology (E-JUST), Alexandria, Egypt*
Islam.Ali@ejust.edu.eg
[3]*Department of Computer Science and Engineering, Egypt-Japan University of Science and Technology (E-JUST), Alexandria, Egypt*
Walid.Gomaa@ejust.edu.eg

## ABSTRACT

Asset health monitoring continues to be of increasing importance on productivity, reliability, and cost reduction. Early Fault detection is a keystone of health management as part of the emerging Prognostics and Health Management (PHM) philosophy. This paper proposes a Hidden Markov Model (HMM) to assess the machine health degradation. using Principal Component Analysis (PCA) to enhance features extracted from vibration signals is considered. The enhanced features capture the second order structure of the data. The experimental results based on a bearing test bed show the plausibility of the proposed method.

## 1. INTRODUCTION

Condition-Based Maintenance (CBM) strategy is preferred over preventive maintenance for critical equipment. It is established that CBM reduces downtime and maintenance costs while it increases systems reliability [1], [2]. The well-established PHM philosophy extends CBM to optimize maintenance by further enhancing reliability, life expectancy, detection of degradation, and operational availability of systems while, at the same time, further decreasing Life-Cycle costs and downtime, in addition to estimating the remaining useful life of systems [3], [4]. This philosophy uses available real-time sensory data obtained from equipment and is defined as an approach for the health management of systems based primarily on fault detection, diagnostics, prognostics, and maintenance decision-making [5], [6].

Fault detection oversees identifying whether the monitored component or process is properly working or not. It is the stepping-stone for diagnostics and prognostics. It requires gathering data for normal operating conditions to compare them with actual operation. Normal operating conditions data are abundant. Hence, fault detection using data-driven models is within reach compared to diagnosis and prognosis that require gathering data about faulty conditions in highly reliable and critical systems, which is found to be difficult [7].

A review of data-driven Fault Detection and Diagnosis (FDD) methods is provided in [8], covering the different data-driven FDD frameworks, including machine learning, signal-based, and knowledge-based methods. The review also summarizes the challenges of data-driven FDD field implementations. The review notes HMMs as one of the models adopted for multimode process monitoring due to their robust stochastic and inferential features. HMMs are considered a promising trend for research and are exploited for fault detection in actual conditions where readings of multiple physical measurements available are sparse and asynchronous in [7]. Other examples of applying HMM in FDD are presented in [9]–[17].

HMMs performance is usually improved by training it on useful features processed from the original signal. Principal Component Analysis (PCA) can be used in several ways for features extraction in combination with HMM. A recent work on an HMM-based fault detection system depicted in [7], employed PCA for dimension reduction through projecting the variables in a new space and selecting the principal components. [18] applies Dynamic PCA (DPCA), an extension of PCA, to extract features that consider serial correlations in signal processed data.

PCA is also used in other applications for feature extraction such as electroencephalogram (EEG) in the domain of Brain-Computer Interface (BCI). [19] generates autoregressive (AR) parameters from EEG data and uses PCA to decrease

---

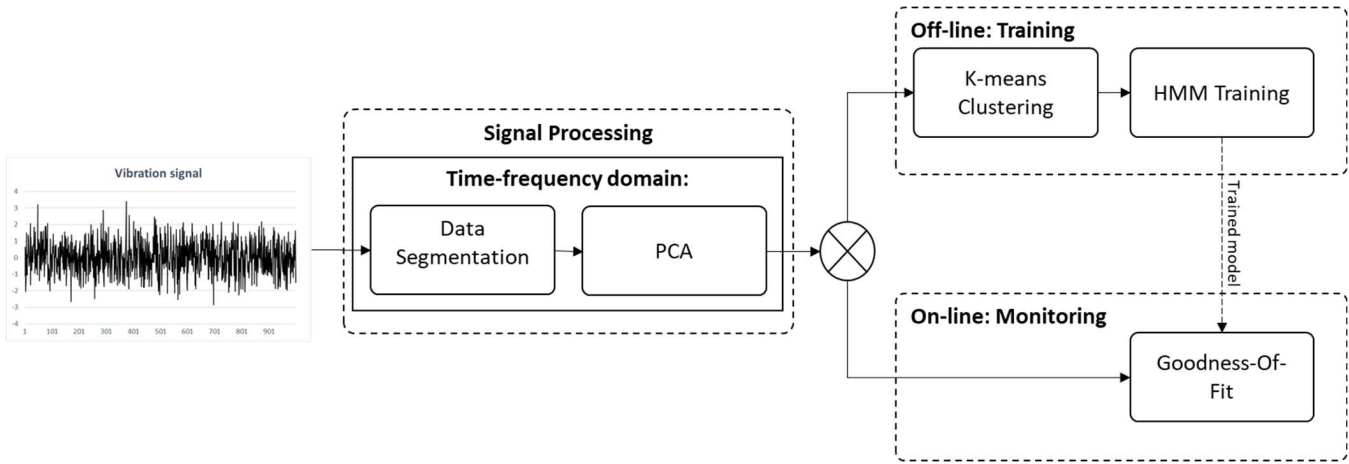

**Figure 1 Bearing fault detection framework**

the number of features. In [20], PCA was used to find principal component features that capture the second-order statistical structure of the data in a manner similar to wavelet transform.

Rolling bearings are of great importance to all forms of rotating machinery and are among the most common machine elements. Its failure is one of the leading causes of breakdowns in rotating machinery and can be catastrophic [21]. Vibration analysis is widely used in fault detection and bearings diagnosis [18], [22].

In this paper, an HMM fault detection method is trained by principal component features that capture the second-order statistical structure of rolling bearings vibration signals which was initially proposed for EEG data [20]. The results show that using PCA for vibration signal processing yields good results when used for fault detection using HMMs.

The remainder of this paper is organized as follows: Section 2 presents the proposed data-driven fault detection approach; Section 3 details the PCA signal processing technique; Section 4 introduces the HMM model; Section 5 presents the results obtained from applying the proposed method to Intelligent Maintenance Systems (IMS) bearing data; and finally, Section 6 concludes the paper.

## 2. PROPOSITION

This study proposes a bearing fault detection model by jointly applying PCA and HMM methods. PCA is used for signal processing because of its ability to capture the second-order statistical structure of the data. Main principal components from the PCA are used to train the HMM. K-mean clustering is applied to the training data and optimum number of clusters is used as the number of HMM states. Baum-Welch algorithm is used to train the HMM and the forward-backward algorithm is used to estimate the samples Log-Likelihood (LL). Online Health degradation is monitored by applying goodness-of-fit test to the monitored sample. The Z-score is computed for the monitored/ test sample and compared to the training sample distribution. Figure 1 sketches the proposed framework.

## 3. SIGNAL PROCESSING

Rolling bearings failure characteristics, together with their vibration measurement and evaluation, have been thoroughly studied and standardized [23], [24]. Bearings generally go through four stages of fault progression to failure [25]–[27]. Stage 1 is a very early stage of failure indicating slight defects detectable only in the ultrasonic frequency range >20KHz. In stage 2, the defects become larger ringing the natural frequencies of the bearing. This stage occurs in the 500-2000Hz frequency range. In stage 3, defects and wear become visible and the bearing defect frequencies start to appear in the velocity spectrum. Finally, in stage 4 the natural frequencies and some of the bearing defect frequencies decrease in amplitude and just prior to total failure the amplitudes in the highest frequency region defined in stage 1 may grow excessively. Bearing stages of damages are depicted in Figure 2.

Vibration signal processing and features extraction have been investigated. Usually, signals are processed in the time, frequency, and time-frequency domains. Time-domain features are usually sensitive to impulsive oscillations and frequency domain features attempt to find the characteristics frequencies related to the rotation of bearings like the Ball-Pass Frequency of Outer ring (BPFO) and Ball-Pass Frequency of Inner ring (BPFI) for bearing health degradation monitoring.

While the frequency domain analysis averages transient vibrations, the time-frequency domain analysis resolves this issue. It is used to show how frequency changes with time. Wavelet transform is usually the tool of choice for time-frequency domain analysis.

In this paper, PCA for time-frequency domain analysis of vibration signals is considered. PCA resembles wavelet transform in learning basis functions. However, they have the difference that PCA learns basis functions from the ensemble of data, while wavelet transform uses basis functions that are fixed in advance [20]. This method was developed for and tested on EEG data in the BCI domain.

The method starts by employing the data segmentation procedure where the vibration signals are broken down into overlapping segments. Principal component features are then extracted from each segment.

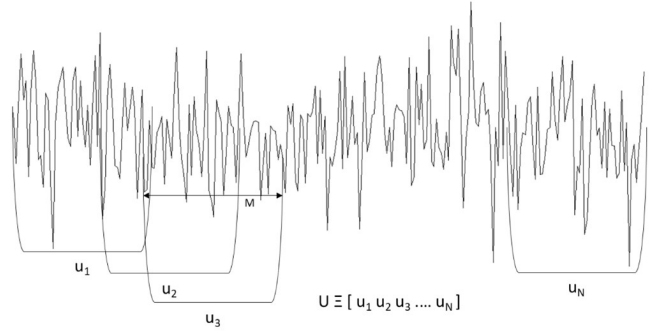

Figure 3 Overlapping windows grouped in matrix U

### 3.2. PCA

PCA is mainly a dimensionality-reduction method. It applies orthogonal linear transform to map the data to a new space, called eigenspace, such that maximum variance is retained in the main components [29], [30].

PCA transforms u to v = Wu (where u is the observation vector). Eigen decomposition is used to calculate eigenvalues and eigenvectors. Letting the data covariance matrix be $R_u$ in which

$$R_u = U_u D_u U_u^T$$

Where $U_u$ is the eigenvector matrix of $R_u$ and $D_u$ is the corresponding diagonal matrix of eigenvalues. Hence, letting $W = U_u^T$, transforming the data is achieved by

$$v = W u$$

Dimension reduction is achieved by ordering eigenvalues and selecting the corresponding p columns of $U_u$ where p < M. W is reconstructed to the p x M matrix of the p dominant column vectors in $U_u$.

PCA is applied to the decomposed time series matrix U = M x N (see Figure 3) and is used it to find a smaller p by M matrix W for PCA. In the case of this study, W matrix is calculated for each of the directions where vibration was measured, i.e., $W_x$ and $W_y$. (X and Y denoting vibration measurement directions) for the tests conducted with measurements in X and Y directions. The principal component features vector for the vibration signal in each direction is then computed by $v_n = W_{Vn}u_n$ (where n stands for X or Y directions). Principal component features extracted from each direction are then concatenated with each other. Useful information is expected to be retained in the principal directions. Therefore, it is expected that the PCA will extract useful features from the vibration signals [20].

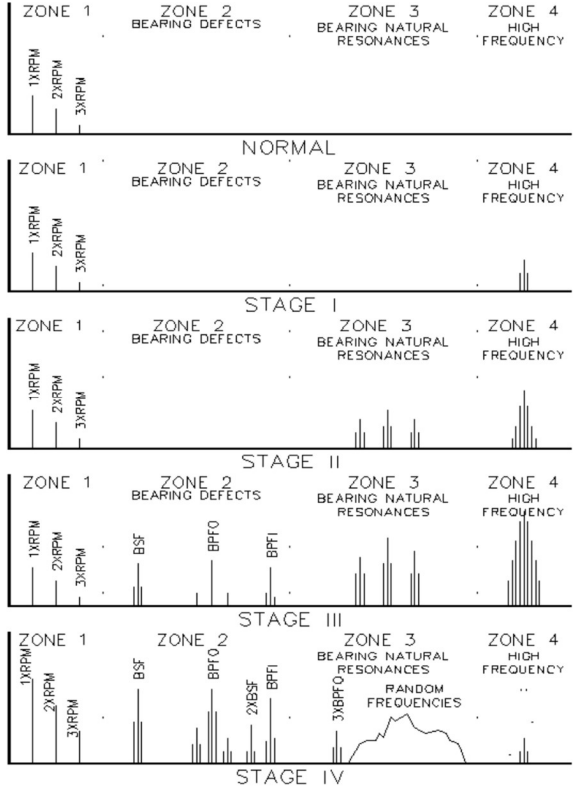

Figure 2 Bearing failure stages [28]

### 3.1. Data segmentation

Vibration signal in each direction is decomposed into overlapping blocks where principal components are extracted from each block. The principal components will carry useful information and retain maximum variance, while the slight noise will appear in the minor components.
Time-series is decomposed into N overlapping blocks. The blocks are used to construct an M x N data matrix (where M is the length of time of the data block as per). The block window length and the overlap length are parameters to be optimized in the model. The block window and overlap lengths should maintain M remaining less than N (M<N).

### 4. HMM

HMMs are dynamic classifiers used in a variety of fields, most notably in the field of speech recognition [31]. HMMs are an extension of Markov chains. They are made of a defined number of states; each state generates an observation at a specified time point. The HMM sequence of states is hidden and not directly observable. It is characterized by the following elements:

1- **S**, A defined number of states;
2- **A**, the state transition probability matrix;
3- **O**, the number of distinct observation symbols per state;
4- **B**, the observation symbol probability matrix;
5- **π**, the initial state distribution.

Given observation sequences $O = \{O_1, O_2, O_3...O_T\}$, a complete specification of an HMM requires specifying the model parameters $\lambda = \{A, B, \pi\}$, state sequence $S = \{S_1, S_2, S_3...S_T\}$, and the probability distribution of the observation sequence given a model $P(O|\lambda)$. Three basic algorithms are used to solve the three main issues of HMM:

1- Baum–Welch algorithm: This algorithm solves the problem of HMM training. This algorithm is used to find the model parameters $\lambda = \{A, B, \pi\}$ that maximize the probability $P(O | \lambda)$.
2- Forward–backward procedure: Given an HMM model $\lambda = \{A, B, \pi\}$ and input observation sequence $O = \{O_1, O_2, O_3...O_T\}$, this algorithm computes $P(O | \lambda)$ that the model estimates.
3- Viterbi algorithm: With the trained HMM model λ, given an observation sequence O, this algorithm finds a corresponding state sequence $S = \{S_1, S_2, S_3...S_T\}$, which best explains the observations.

### 4.1. Offline: Training

The Baum-Welch algorithm is used to train the corresponding HMM fed with the concatenated features from the data segmented PCA, the model is trained on normal condition data. The Log-Likelihood (LL) of individual training samples was reached using the forward-backward algorithm.

### 4.2. Online: Health Degradation Monitoring

The health degradation is monitored by computing $P(O | \lambda)$ representing the likelihood of the monitored observation sequence given the trained model. This is achieved by applying a special form of the forward-backward procedure, the forward algorithm. The forward-only variant sums the forward recursion algorithm run on the entire sample observation sequence for all states and returns the LL of the monitored sample [32]. Then, goodness-of-fit test between the training samples' likelihood distribution and the monitored sample is executed. A fault is detected when the monitored sample's log-likelihood deviates from the training samples' distribution [7].

## 5. EXPERIMENTAL RESULTS

### 5.1. Setup and Data

Bearing run-to-failure tests were performed under normal load conditions on a specially designed test rig, shown in Figure 4. This test overcomes the lower capability to discover natural defect propagation in the early stages in experiments using defective bearings [21].

Four double row bearings (Rexnord ZA-2115) were mounted on one shaft. The rotation speed was constant at 2000 RPM. The Shaft was rotated by an AC motor coupled to the shaft through rub belts. A radial load of 6000 lbs is applied onto the shaft and bearing by a spring mechanism. All bearings are force lubricated.

A PCB 353B33 High Sensitivity Quartz ICPs Accelerometer was installed on each bearing housing. Vibration data were collected every 20 minutes by a National Instruments DAQCard-6062E data acquisition card. The data sampling rate is 20 kHz, and the data length is 20480 points per sample.

Prognostics Center of Excellence shared through a prognostic data repository three data sets (sets 1, 2, and 3), contributed by the Center for IMS – University of Cincinnati, OH, that were generated from the setup. Each data set describes a test-to-failure experiment where four new bearings were installed, keeping other experimental conditions the same. Each data set consists of individual files that are ~1-second vibration signal snapshots recorded at specific intervals.

Bearing degradation is monitored through a magnetic plug installed in the oil feedback pipe that collects debris from the oil. The oil circulation system regulates the lubricant's flow and temperature. The test is stopped when the accumulated debris exceeds a preset level, leading the switch to turn off.

The full life data from two representative bearings (bearing 2 and bearing 3 from set 1), whose faults include inner race defect of bearing 3, are used to test the performance of the proposed method. More details about each experiment and accompanied failures are described in [33].

### 5.2. Offline Training

The first one--third of the bearing's whole life is chosen as the healthy data set used for training. From this portion of the data, it is assumed the first one fifth to be early-stage operation, and is excluded from the training data. Chosen healthy state readings from Bearing 2 and Bearing 3 are further detailed in Table 1.

**Table 1 Healthy state reading used for training**

|  | Bearing 2 | Bearing 3 |
|---|---|---|
| Readings | 2156 | 2156 |
| 1/3rd the readings | Readings 1 - 719 | Readings 1 - 719 |
| Omit early 20% | Readings 144 - 719 | Readings 144 - 719 |

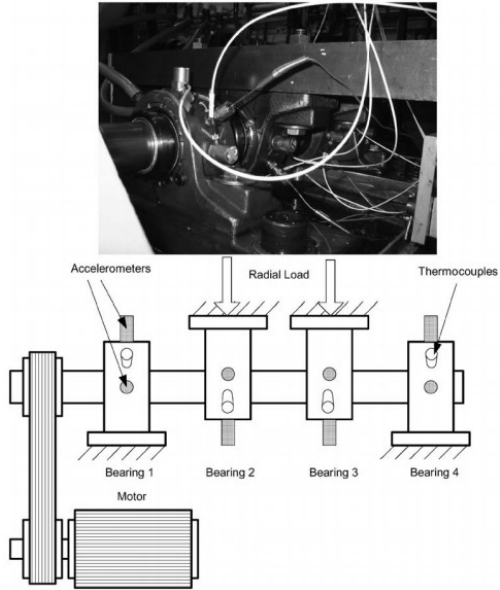

**Figure 4 Bearing test rig and sensors placement [20]**

Signal processing is initiated by applying data segmentation. A window size of 512 readings and overlap every 32 readings is chosen. This choice ensures the window size M remains smaller than the total number of overlapping windows. The PCA is run next and the first ten eigenvectors corresponding to the largest ten eigen values i.e., setting $p = 10$, are kept. It is worth noting that the first 2 Eigenvalues were always retaining more variance, but due to the large size of variables, we chose ten principal components to represent each window. The ten main components retain about 20% of the vibration signal variance in each direction and compress the original vibration signal at the same time.

The HMM is then trained on the selected main components for each bearing experiment. The training data is clustered and the HMM number of states is set to equal the number of clusters. K-means clustering that exploits the Euclidean metric is used. Accordingly, the data is clustered by minimizing the Euclidean intra-cluster distance and maximizing the inter-cluster Euclidean distance [7]. The elbow method [34], [35] is applied to choose the optimal number of clusters. Figure 5 and Figure 6 show the elbow graph for bearing 2 and bearing 3, respectively. Bearing 2 clusters are circled to four and Bearing 3 to three.

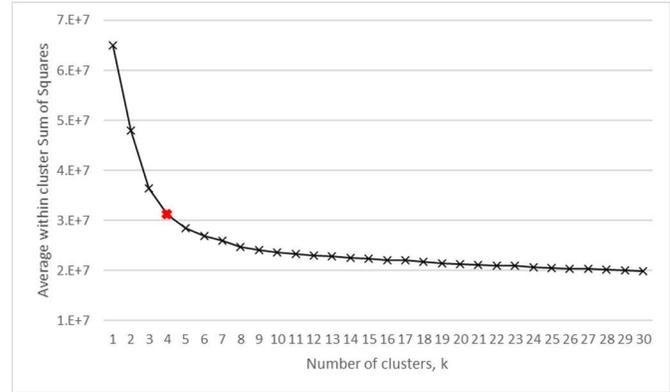

**Figure 5 Bearing 2 circling optimal k at 4 clusters**

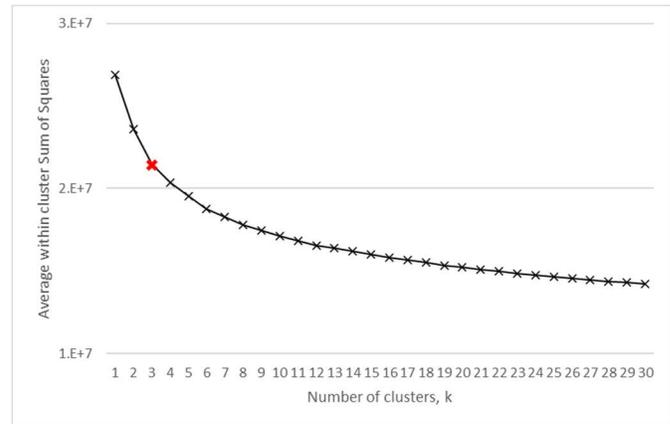

**Figure 6 Bearing 3 circling optimal k at 3 clusters**

### 5.3. Online Degradation Monitoring

After building the model, fault detection should follow. Applying the Shapiro-Wilk test [36] to the LLs of the training samples and inspecting their corresponding density plot, with the LL of each measurement being independent, show the LLs are approximately normal. Therefore, given the large training sample, the Z-score is calculated for the samples representing the entire life of the bearing to assess each sample's goodness-of-fit. As per Figure 1 in the proposition, the monitored/ test sample passes through the same data processing that the training data went through prior reaching the goodness-of-fit test. The number of states was identified by the aid of the k-means clustering in the training phase. The monitored/ test sequence is fed to the trained model, the LL is returned, and Z-score estimated to verify if the sequence fits the trained model distribution, i.e., bearing is still operating normally. A left side negative Z-score of 2.33 is set as the threshold to trigger alarms that indicate whether a bearing is starting to experience health degradation. The Z-score results based on PCA signal processing and HMM are presented in Figure 8 and Figure 7 for Bearing 2 and Bearing 3 runs to failure, respectively.

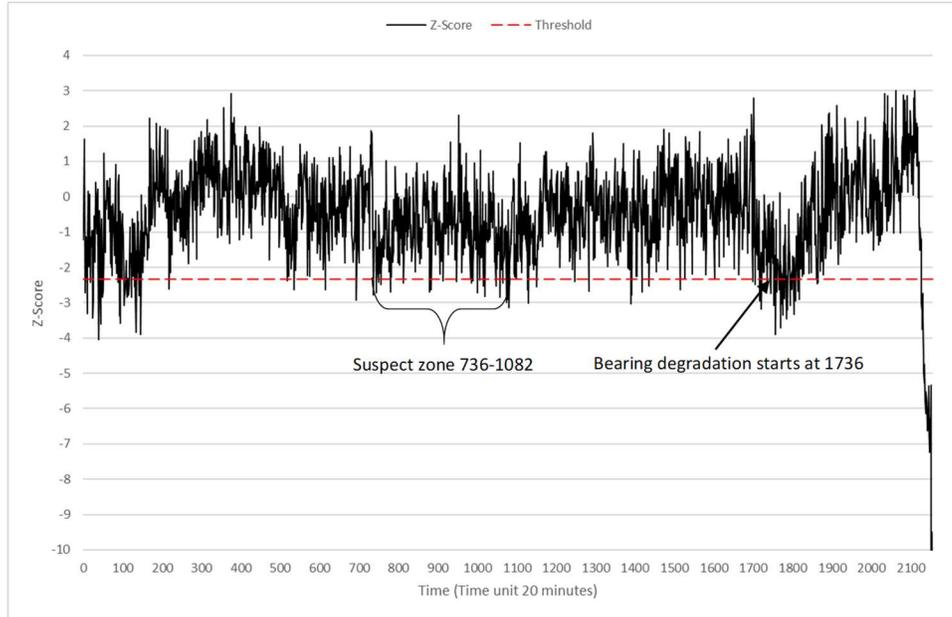

**Figure 8 Health monitoring for the full lifetime of Bearing 2**

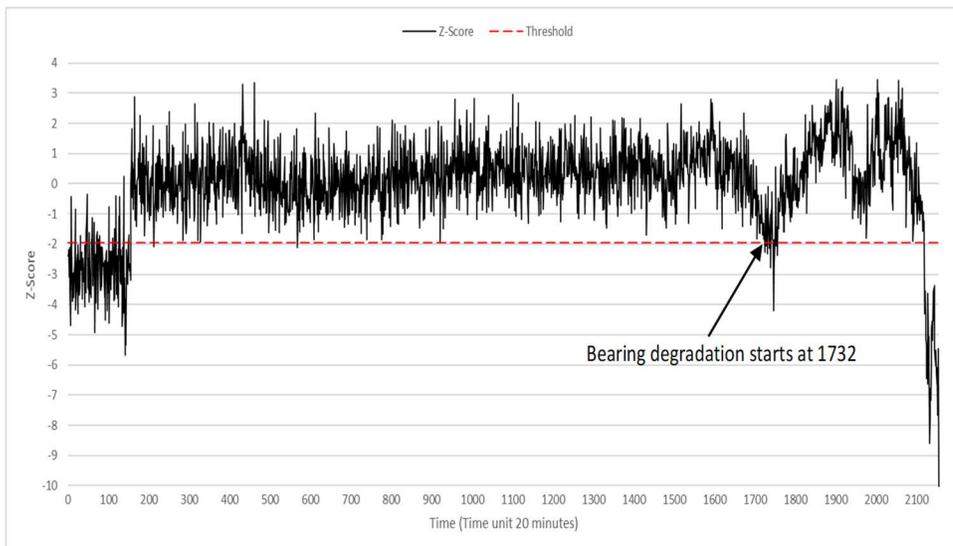

**Figure 7 Health monitoring for the full lifetime of Bearing 3**

Figure 8 shows expected early-stage alarms and, in addition, the bearing was suspect during its mid-life operation before showing some self-healing, yet with few outliers, effects. It also shows that bearing 2 starts experiencing degradation at measurement 1736 before progressing to stage IV of bearing failure - end of life - followed by imminent failure at reading 2127.

Like bearing 2, Figure 7 shows expected early-stage alarms for bearing 3. However, the bearing undergoes much smoother operation till it starts experiencing degradation at measurement 1732 followed by reaching stage IV of bearing failure and inner race failure at reading 2120.

**6. CONCLUSION**

a health monitoring framework for bearings that jointly employs PCA and HMM was presented. The PCA method used for signal processing is originally proposed for EEG data in the BCI domain and the experimental study on the IMS bearing data demonstrated it is good feature extractor for vibration data. We are currently further investigating its performance, comparing it to other signal processing techniques, especially wavelet transform, in addition to using it in tandem with signal processing in the time and frequency domains.

ACKNOWLEDGEMENT

This work was supported by computational resources provided by the Bibliotheca Alexandrina (hpc.bibalex.org) on its High-Performance Computing (HPC) infrastructure.

We thank Prof. M.A Younes (Faculty of Engineering, Alexandria University, Alexandria, Egypt) for his valuable inputs on handling experiment's results.